%% file: acl_latex.tex
\pdfoutput=1

\documentclass[11pt]{article}

\usepackage[preprint]{acl}

\usepackage{times}
\usepackage{tcolorbox}
\usepackage{latexsym}
\usepackage{multicol}
\usepackage{multirow}
\usepackage{booktabs}
\usepackage{microtype}
\usepackage{graphicx}   
\usepackage{subcaption} 
\usepackage[T1]{fontenc}

\usepackage[utf8]{inputenc}

\usepackage{microtype}

\usepackage{inconsolata}

\usepackage{graphicx}

\title{ReadBench: Measuring the Dense Text Visual Reading Ability of Vision-Language Models}

\author{Benjamin Clavi\'{e} \\
  Answer.AI \\
  Japan \\
  \texttt{bc@answer.ai} \\\And
  {Florian Brand} \\
  Artificial Intelligence and Intelligent\\ Information Systems, University of Trier \\
  German Research Center for\\ Artificial Intelligence (DFKI), Trier
  \\ \texttt{brand@uni-trier.de,florian.brand@dfki.de}
}

\begin{document}
\maketitle
\begin{abstract}
Recent advancements in Large Vision-Language Models (VLMs), have greatly enhanced their capability to jointly process text and images. However, despite extensive benchmarks evaluating visual comprehension (e.g., diagrams, color schemes, OCR tasks...), there is limited assessment of VLMs' ability to read and reason about text-rich images effectively. To fill this gap, we introduce ReadBench, a multimodal benchmark specifically designed to evaluate the reading comprehension capabilities of VLMs. ReadBench transposes contexts from established text-only benchmarks into images of text while keeping textual prompts and questions intact. Evaluating leading VLMs with ReadBench, we find minimal-but-present performance degradation on short, text-image inputs, while performance sharply declines for longer, multi-page contexts. Our experiments further reveal that text resolution has negligible effects on multimodal performance.
These findings highlight needed improvements in VLMs, particularly their reasoning over visually presented extensive textual content, a capability critical for practical applications.
ReadBench is available at \href{https://github.com/answerdotai/ReadBench}{this https url}.
\end{abstract}

\section{Introduction}

Recently, Large Language Models (LLMs) have rapidly progressed towards multimodality, handling inputs beyond pure text~\cite{multimodality}. In particular, Vision-Language Models (VLMs), which jointly process text and images, have experienced significant improvements starting from early alignment models like CLIP~\cite{clip} and ALIGN~\cite{align} to recent "frontier" models such as GPT-4o~\cite{gpt4o} and Gemini 1.5~\cite{gemini1.5}, as well as leading open-weight models including Pixtral~\cite{pixtral} and Qwen2.5-VL~\cite{qwen2.5-vl}.

Numerous benchmarks have been introduced to measure and support the rapid development of these models. Most focus primarily on visual comprehension tasks (MMMU~\cite{mmmu}, DocVQA~\cite{docvqa}), assessing a model's ability to leverage visual elements. Other benchmarks emphasize transcription accuracy (OCRBench~\cite{ocrbench}) or reasoning about visual scenarios containing text, such as pictures of street signs (TextVQA~\cite{textvqa}).

A third category of benchmarks, examplified by ViDoRe~\cite{colpali}, assesses multimodal retrieval, i.e. the ability to retrieve relevant visual documents given a textual query. 
Multimodal retrieval methods, having proven themselves to be superior to text-only approaches~\cite{dse,colpali, ohr}, are garnering increasing interest from both the academic community and industry. This momentum has subsequently fueled enthusiasm for Visual Retrieval-Augmented Generation~\cite{rag} (VisRAG~\cite{visrag}).

While they do contain visual information, such as graphs or tables, real-world documents are generally text-heavy~\cite{due}, requiring the models to be able to achieve strong "reading" performance to generate satisfying answers. This capability is currently inadequately assessed by common benchmarks, which either primarily focus on non-textual visual understanding, or, as is the case for transcription-focused benchmarks, require little understanding ability. Some effort toward better evaluating reading capabilities were introduced in PixelWorld~\cite{pixelworld}, albeit a focus on single-modality inputs. However, to represent real world usage, it is crucial that these capabilities be measured in a truly multi-modal fashion, where text instructions are given alongside image documents. 

\textbf{Contribution}
In this paper, we introduce ReadBench, a benchmark designed specifically to evaluate VLMs' ability to effectively answer questions based on text presented visually. ReadBench adapts widely-used textual benchmarks into multimodal format, converting contexts into images while preserving textual prompts and questions. Evaluating state-of-the-art VLMs, we observe universal performance degradation in multimodal versus purely textual scenarios. While performance degradation varies across models, the general trend shows that it is minimal for short inputs but significant for multi-page contexts, often causing double-digit percentage drops. Interestingly, we find substantial variation across models in problematic inputs, suggesting model-specific rather than universal challenges. Finally, ablations reveal negligible effects from input resolution, reinforcing previous findings that VLM performance remains stable even with low-resolution textual images~\cite{giffmanathelowresherald}.

\section{Constructing ReadBench}

ReadBench aims to evaluate how effectively VLMs read and extract textual information from images, while answering a question in textual form, mimicking realistic VisRAG scenarios~\cite{visrag,hfmmrag}, where successfully following a task given through textual instructions require understanding visually-presented context.

To achieve this, we convert widely-used text-only benchmarks into mixed-modality inputs: instructions and questions remain textual, while contexts are presented as images of text. 

\subsection{Makeup}
\begin{figure}[]
    \centering
    \includegraphics[width=0.9\linewidth]{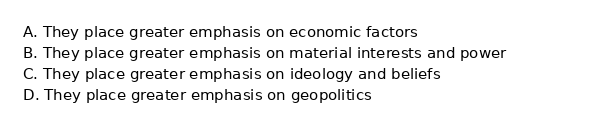}
    \caption{An example MMLU-Redux converted input}
    \label{fig:reduxexample}
\end{figure}

We sample from five standard permissively licensed evaluation benchmarks to assess various VLM reading capabilities: three common short-context benchmarks and two long-context evaluation suites. For the long-context benchmarks, we select a maximum token count bin of 8k (averaging to 12 pages of text in image format). This decision reflects the fact that VLM performance already reaches significant degradation at that stage, while also addressing practical issues by reducing evaluation costs. We only include English-language inputs.

\begin{figure}[!h]
    \centering
    \includegraphics[width=0.95\linewidth]{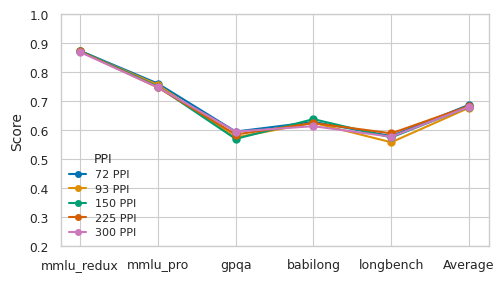}
    \caption{Gemini 2.0 Flash multi-modal scores across a range of PPI settings (resulting in different resolutions).}
    \label{fig:resolution}
\end{figure}

\subsubsection{Short-Context}
\textbf{MMLU-Redux}~\cite{mmluredux}, an updated version of MMLU~\cite{mmlu} improving the overall quality of the dataset, notably by removing ambiguous questions.\\
\textbf{MMLU-Pro}~\cite{mmlupro}, a harder variant focusing on STEM, with most questions having 10 possible answers rather than 4. \\
\textbf{GPQA-Diamond}~\cite{gpqa},
another multiple-choice question dataset focusing on difficult graduate-level science questions.

\subsubsection{Long-Context}

\begin{figure*}[!h]
        \centering
        \includegraphics[width=0.9\linewidth]{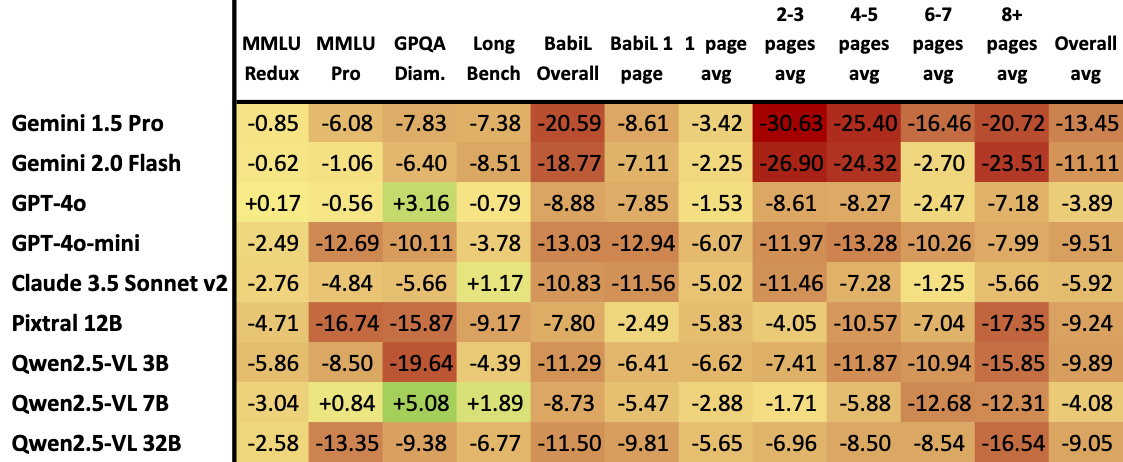}
        \caption{Performance degradation overview across datasets for all models}
        \label{fig:main}
    \end{figure*}

\textbf{BABILong} evaluates the ability of models to reason over simple facts contained within various lengths of inputs. We sample from all ten questions it contains, the first of which is a Needle-in-a-Haystack (NIAH) task~\cite{niah}, simply requiring extracting a factoid from the context without any reasoning. All other questions introduce mildly more complex reasoning tasks, such as counting, connecting related facts, or handling chained negations. Despite its simplicity, it has proven challenging for all frontier models, thus making more complex benchmarks currently unnecessary~\cite{babilong}, a result supported by recent work~\cite{nolima}. \\
\textbf{LongBench} We specifically select 4 subsets of the LongBench~\cite{longbench} benchmark representing common QA datasets: HotPotQA~\cite{hotpotqa}, NarrativeQA~\cite{NarrativeQA}, TriviaQA~\cite{triviaqa} and 2WikiMultiHopQA~\cite{wikimqA}.
To ensure the results can be easily compared to the rest of our chosen datasets, we use a simplified binary metric\footnote{We discard the original, more fine-grained metrics} where overlap between the gold standard and the model answer is considered a positive label.

\subsection{Image Generation Process}
\label{sec:ReadBench}
    
To closely mimic real-world document scenarios, ReadBench generates images based on the widely-used A4 paper format\footnote{Roughly equivalent to the US Letter format}. Following common practices and the W3C's accessibility and readability recommendations~\cite{w3c}, we set font size to 12 (Arial) and adopt a 92.9 Pixels-Per-Inch (PPI) ratio to determine the resolution, closely matching common screen resolutions and VLM input constraints~\cite{gemini1.5,windowsppi}. 

For short-context benchmarks, we keep instructions and questions as text, converting only answer options to images. For long-context QA datasets, instructions and questions are kept in textual format while the entire context needed to answer the questions are converted to images. We choose this strategy to more accurately mimic common real-world usage of multimodal inputs, where textual questions are asked about visual documents~\cite{hfmmrag,visrag}.

Once a page is full, a new page is created, until the entire context has been converted. Partially filled pages are cropped to remove excessive whitespace. An example generated image for MMLU-Redux is shown in Figure~\ref{fig:reduxexample}.

\subsubsection{Impact of Resolution}
\label{sec:resolution}

It has previously been noted that lower resolution, or "blurry images", have little impact on the performance of vision models~\cite{giffmanathelowresherald}. Before settling on our final 92.9PPI ratio, we conducted evaluations with multiple PPI settings with Gemini 2.0 Flash. We evaluate a range of PPI values, from 72PPI, a common low-resolution setting, to 300PPI, frequently associated with "retina" high-quality resolutions~\cite{ppiref}.

Our results, presented in Figure~\ref{fig:resolution}, reveal little variation in model performance across resolutions, confirming that lower-resolution inputs do not negatively affect VLM reading performance.

\subsection{Sampling Strategy}

Evaluating VLMs on image-based inputs can quickly become prohibitively expensive, especially as processing image input has the potential to lead to larger token counts than textual inputs~\cite{qwen2.5-vl,paligemma}. To increase benchmarking efficiency, we randomly sample up to 35 examples per benchmark subset, a size we experimentally found to yield results highly correlated with evaluating the full dataset.

\section{Benchmarking Results}

\begin{figure*}[!th]                
  \centering
  \begin{subfigure}[t]{0.42\textwidth}
        \centering
    \includegraphics[width=\linewidth]{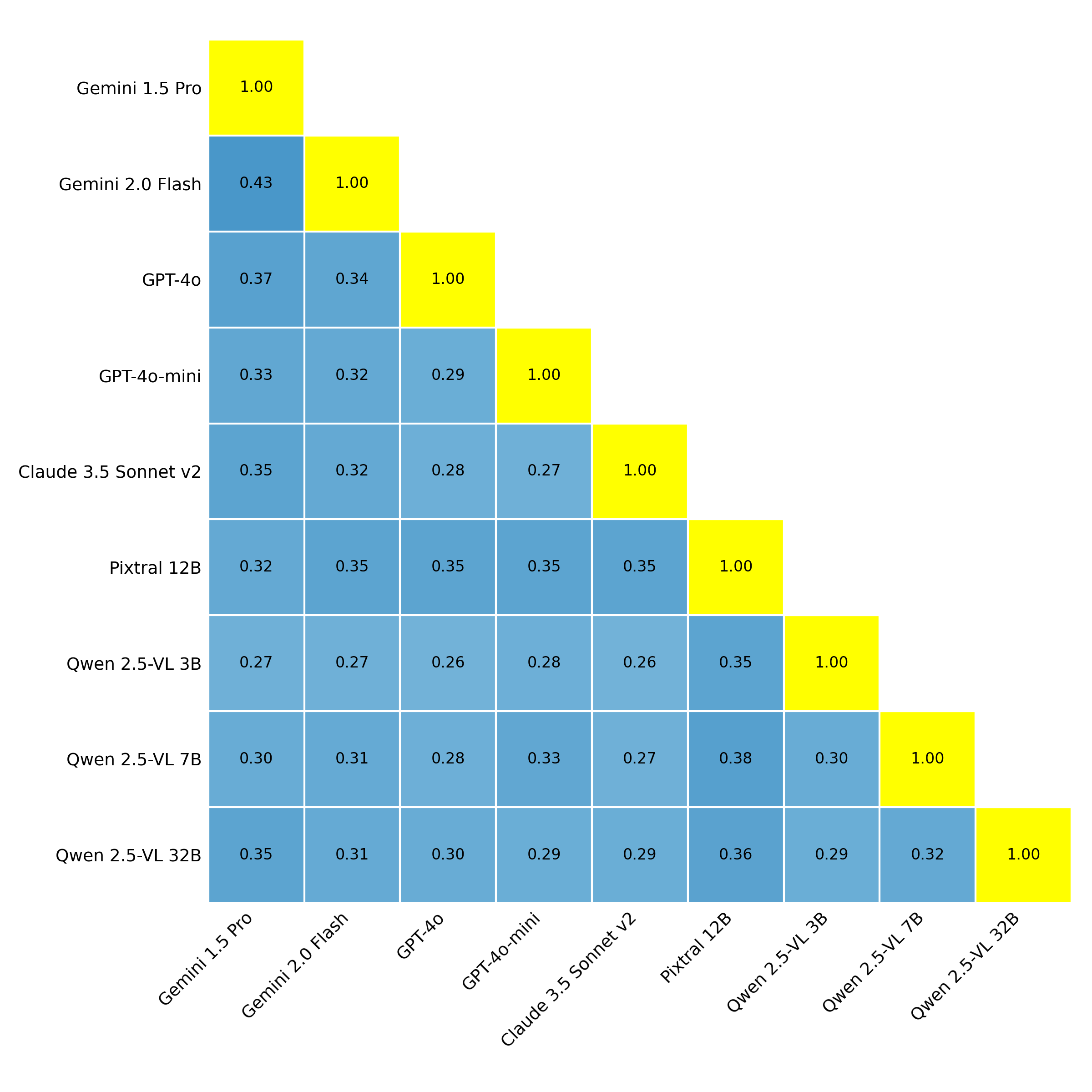}
    \caption{Jaccard similarity of questions with multimodal/text mismatches across models.}    
    \label{fig:jaccard}
  \end{subfigure}
  \hfill                    
  \begin{subfigure}[t]{0.5\textwidth}
    \centering
    \includegraphics[width=\linewidth]{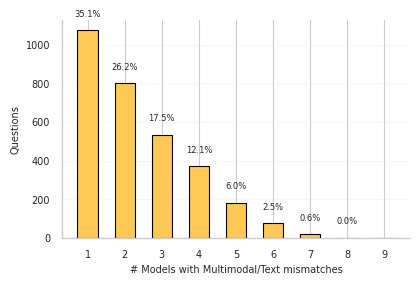}
    \caption{Distribution of the number of models yielding multimodal/text mismatches for each question.}  
    \label{fig:mismatches}
  \end{subfigure}

  \caption{Consistency of multimodal–text disagreements across models.}
  \label{fig:sim}
\end{figure*}

We evaluate a selection of current-generation, commonly used VLMs, targeting multiple model sizes, and select both leading open-source models and frontier models from closed-source labs. 

The summary of our evaluation results is presented in Figure~\ref{fig:main}. All results are measured as the average of 3 individual runs to reduce variance.
Overall, we observe universal performance degradation across models when reading textual information from images rather than purely textual inputs. Two main factors appear particularly influential: input length and task difficulty.

\textbf{Input Length} Performance on short contexts (up to one page of text) varies significantly across models. While some models, such as Pixtral 12B and Qwen2.5-VL 32B, experience notable degradation even at shorter lengths, others, such as GPT-4o and Qwen2.5-VL 7B, maintain or slightly improve over their textual performance. However, all models exhibit significant performance drops when contexts span multiple pages, with Gemini 1.5 Pro experiencing the largest relative degradation (around 30\%) on inputs of 2–3 pages while GPT-4o fares the best out of all evaluated models.

\textbf{Task Difficulty} The short-context multiple-choice benchmarks we selected, 
represent three levels of difficulty, with MMLU-Redux being the easiest and GPQA the hardest. We observe a clear correlation between benchmark difficulty and performance degradation: models generally show minimal degradation on the easiest task, but higher variations begin to appear on harder ones. Some outliers exist, with GPT-4o and Qwen2.5-VL 7B reaching stronger results on multimodal GPQA\footnote{This aligns with prior observations that GPT-4o's textual GPQA accuracy declined following its initial release, while its multimodal capabilities improved~\cite{4oweird}.}.

\section{Mismatch Analysis}

We next analyze whether different models fail on similar examples when comparing multimodal to textual inputs. A \textbf{mismatch} occurs when the assessment of the model's answer (correct or incorrect) is different between textual and multimodal settings.

Figure~\ref{fig:jaccard} shows pairwise Jaccard similarities between mismatched question sets across models. Across all model pairs, overlap remains moderate (20-35\%), even within similar model families, indicating limited consistency in problematic inputs.

Figure~\ref{fig:mismatches} presents how often each individual sample triggers mismatches across multiple models. Most questions cause mismatches in only one or two models, while very few (3.1\%) affect more than five models. No single question universally causes mismatches across all models.

These findings suggest that performance degradation primarily arises from model-specific challenges rather than universally problematic inputs.

\section{Conclusion}

In this paper, we introduced ReadBench, a multimodal benchmark designed to evaluate Vision-Language Models' (VLMs) capability to read and extract information from visually-presented textual contexts. ReadBench directly addresses the current lack of benchmarks explicitly testing VLMs on text-rich images, an increasingly relevant scenario given the growth of multimodal applications.

We demonstrate that all state-of-the-art VLMs experience some degree of performance degradation on multimodal inputs compared to purely textual scenarios. While most models handle short contexts reasonably well, significant degradation is observed when processing multi-page visual contexts. Inputs causing degradation across modalities vary substantially across models, suggesting there is no "universal trigger" for this phenomenon. Our ablations also confirm that lower resolution images do not impact model performance.

These insights highlight an important gap in mixed-modality capabilities. As multimodal applications grow in popularity, addressing these challenges will be important for future developments. 

\section*{Acknowledgements}

The authors thank Johno Whittaker for his infinite wisdom on visual models and helpful advice, as well as Kerem Turgutlu for his assistance in running the evaluations in a computationally efficient way. We also thank Aamir Shakir for sharing their work on a benchmark exploring similar problems within visual retrieval pipelines.

\textbf{AI Assistants} Generative AI tools, specifically Anthropic's Claude Sonnet 3.5, OpenAI's o3 and Google's Gemini Pro 2.5, were used in the making of this work, at various stages of the implementations. They also greatly contributed to the cleanup of the code repository, to create an easily shareable and reproduceable version. GPT 4.5 was used as a proof-reader for the paper, contributing minor phrasing changes and typo fixes.

\section*{Limitations}

\textbf{Multilingual} A major limitation of ReadBench is its focus on the English language. While we do plan to explore multilinguality in future work, the vast quantity of resources available in English greatly facilitated the initial development of the benchmark. Multimodal evaluations in non-English language are currently sparse, a problem which our work does not address. We believe future work should extend ReadBench to other languages, such as those supported by MMMLU, a multilingual version of MMLU.

\textbf{Model Selection} We evaluate 9 widely used VLMs on ReadBench, providing an overview of the current performance of state-of-the-art models. However, there exist many more models, which we are unable to exhaustively evaluate due to budget constraints. To mitigate this, we are releasing our full evaluation suite and data, supporting independent evaluations. We hope to be able to support a leaderboard as part of future work.

\textbf{Additional Modalities} Our work largely focuses on exploring multimodality under the Text+Vision angle, specifically focusing on only images. With the recent developments of models handling other kinds of modalities, such as Audio or Video inputs, future work should explore constructing a similar benchmark to evaluate performance across more modalities.

\textbf{Benchmark Simplicity} Our work provides useful insight into the reading abilities of LLMs in different evaluation contexts and shows that there are significant degradations, especially in multi-pages settings. However, it mostly uses standard academic benchmarks, which all have individual flaws, such as BABILong's rather simplistic construction. While these benchmarks are currently useful to evaluate current model capabilities, a future avenue of research would be exploring more complex dataset construction in order to be able to accurately evaluate ever-improving models.

\bibliography{acl_latex}

\appendix

\section{Prompts}
\label{app:prompt}

In all cases, our prompts are largely adapted from those of the original datasets we sample from: MMLU-Pro~\cite{mmlupro}, MMLU-Redux~\cite{mmluredux}, BABILong~\cite{babilong} and LongBench~\cite{longbench}, with mild modifications to ensure multimodal support.

\input{prompt}









\end{document}

%% file: prompt.tex
\subsection{MMLU (Pro and Redux)}

\subsubsection{Text Only}

\begin{tcolorbox}[colback=gray!10, colframe=black!50, title=MMLU (Text-only)]
The following are multiple-choice questions (with answers) about \texttt{\$CATEGORY}. Think step by step and then finish your answer with \texttt{The answer is (X)} where X is the correct letter choice.

\textbf{Question:} \texttt{\$QUESTION}

\textbf{Options:}\\
\texttt{\$OPTIONS}
\end{tcolorbox}

\subsubsection{Multimodal}

\begin{tcolorbox}[colback=gray!10, colframe=black!50, title=MMLU (Multimodal)]
The following are multiple-choice questions about \texttt{\$CATEGORY}. The possible answers are listed in the attached image(s).

\textbf{Question:} \texttt{\$QUESTION}

\textbf{Options:}\\ \textcolor{red}{\{IMAGES HERE\}}

Think step by step and then output the answer in the format \texttt{"The answer is (X)"} at the end, where X is the letter associated with the correct answer.
\end{tcolorbox}

\subsection{GPQA}

\subsubsection{Text Only}

\begin{tcolorbox}[colback=gray!10, colframe=black!50, title=GPQA (Text-only)]
What is the correct answer to this question: \texttt{\$QUESTION}

\textbf{Options:}\\
\texttt{\$OPTIONS}

Your answer must end with \texttt{Answer: (LETTER)}. Let's think step by step:
\end{tcolorbox}

\subsubsection{Multimodal}

\begin{tcolorbox}[colback=gray!10, colframe=black!50, title=GPQA (Multimodal)]
What is the correct answer to this question: \texttt{\$QUESTION}

\textbf{Options:}\\ \textcolor{red}{\{IMAGES HERE\}}

Your answer must end with \texttt{Answer: (LETTER)}. Let's think step by step:
\end{tcolorbox}

\subsection{Babilong}

Babilong contains 10 individual questions. The general templates are listed first, followed by the per-question instructions used to fill in these templates.

\subsubsection{Text Only}

\begin{tcolorbox}[colback=gray!10, colframe=black!50, title=Babilong (Text-only)]
\$PRE\_INSTRUCTION

\textbf{Context:}
\begin{flushleft}
\texttt{\$CONTEXT}
\end{flushleft}

\textbf{Question:}
\begin{flushleft}
\texttt{\$QUESTION}
\end{flushleft}

\$POST\_INSTRUCTION
\end{tcolorbox}

\subsubsection{Multimodal}

\begin{tcolorbox}[colback=gray!10, colframe=black!50, title=Babilong (Multimodal)]
\$PRE\_INSTRUCTION

\textbf{Context:}\\
\textcolor{red}{\{IMAGES HERE\}}

\textbf{Question:}
\begin{flushleft}
\texttt{\$QUESTION}
\end{flushleft}

\$POST\_INSTRUCTION
\end{tcolorbox}

\subsubsection{Per-Question Instructions}

\begin{tcolorbox}[colback=gray!05, colframe=black!40, title=Babilong QA1]
\textbf{Pre-instruction}\\
I will give you context with the facts about positions of different persons hidden in some random text and a question. You need to answer the question based only on the information from the facts. If a person was in different locations, use the latest location to answer the question.

\vspace{0.5em}
\textbf{Post-instruction}\\
Always return your answer in the following format: The most recent location of ’person’ is ’location’. Do not write anything else after that.
\end{tcolorbox}

\begin{tcolorbox}[colback=gray!05, colframe=black!40, title=Babilong QA2]
\textbf{Pre-instruction}\\
I give you context with the facts about locations and actions of different persons hidden in some random text and a question. You need to answer the question based only on the information from the facts.  
If a person got an item in the first location and travelled to the second location the item is also in the second location.  
If a person dropped an item in the first location and moved to the second location the item remains in the first location.

\vspace{0.5em}
\textbf{Post-instruction}\\
Always return your answer in the following format: The ’item’ is in ’location’. Do not write anything else after that.
\end{tcolorbox}

\begin{tcolorbox}[colback=gray!05, colframe=black!40, title=Babilong QA3]
\textbf{Pre-instruction}\\
I give you context with the facts about locations and actions of different persons hidden in some random text and a question. You need to answer the question based only on the information from the facts.  
If a person got an item in the first location and travelled to the second location the item is also in the second location.  
If a person dropped an item in the first location and moved to the second location the item remains in the first location.

\vspace{0.5em}
\textbf{Post-instruction}\\
Always return your answer in the following format: Before the \$location\_1\$ the \$item\$ was in the \$location\_2\$. Do not write anything else after that.
\end{tcolorbox}

\begin{tcolorbox}[colback=gray!05, colframe=black!40, title=Babilong QA4]
\textbf{Pre-instruction}\\
I will give you context with the facts about different people, their location and actions, hidden in some random text and a question. You need to answer the question based only on the information from the facts.

\vspace{0.5em}
\textbf{Post-instruction}\\
Your answer should contain only one word – location. Do not write anything else after that.
\end{tcolorbox}

\begin{tcolorbox}[colback=gray!05, colframe=black!40, title=Babilong QA5]
\textbf{Pre-instruction}\\
I will give you context with the facts about locations and their relations hidden in some random text and a question. You need to answer the question based only on the information from the facts.

\vspace{0.5em}
\textbf{Post-instruction}\\
Your answer should contain only one word. Do not write anything else after that. Do not explain your answer.
\end{tcolorbox}

\begin{tcolorbox}[colback=gray!05, colframe=black!40, title=Babilong QA6]
\textbf{Pre-instruction}\\
I will give you context with the facts about people and their locations hidden in some random text and a question. You need to answer the question based only on the information from the facts. If a person was in different locations, use the latest location the person was in to answer the question.

\vspace{0.5em}
\textbf{Post-instruction}\\
Your answer should contain only one word – $yes$ or $no$. Do not write anything else after that. Do not explain your answer.
\end{tcolorbox}

\begin{tcolorbox}[colback=gray!05, colframe=black!40, title=Babilong QA7]
\textbf{Pre-instruction}\\
I will give you context with the facts about people and objects they carry, hidden in some random text and a question. You need to answer the question based only on the information from the facts.

\vspace{0.5em}
\textbf{Post-instruction}\\
Your answer should contain only one word – none or \$number\_of\_objects\$. Do not write anything else after that. Do not explain your answer.
\end{tcolorbox}

\begin{tcolorbox}[colback=gray!05, colframe=black!40, title=Babilong QA8]
\textbf{Pre-instruction}\\
I will give you context with the facts about people and objects they carry, hidden in some random text and a question. You need to answer the question based only on the information from the facts.

\vspace{0.5em}
\textbf{Post-instruction}\\
Your answer should contain only one or two words: \$nothing\$ or \$object\$ or \$object\_1\$, \$object\_2\$. Do not write anything else. Do not explain your answer.
\end{tcolorbox}

\begin{tcolorbox}[colback=gray!05, colframe=black!40, title=Babilong QA9]
\textbf{Pre-instruction}\\
I will give you context with the facts about people and their locations hidden in some random text and a question. You need to answer the question based only on the information from the facts. If a person was in different locations, use the latest location the person was in to answer the question.

\vspace{0.5em}
\textbf{Post-instruction}\\
Your answer should contain only one word – \$yes\$ or \$no\$. Do not write anything else. Do not explain your answer.
\end{tcolorbox}

\begin{tcolorbox}[colback=gray!05, colframe=black!40, title=Babilong QA10]
\textbf{Pre-instruction}\\
I will give you context with the facts about people and their locations hidden in some random text and a question. You need to answer the question based only on the information from the facts. If a person was in different locations, use the latest location the person was in to answer the question.

\vspace{0.5em}
\textbf{Post-instruction}\\
Your answer should contain only one word – \$yes\$, \$no\$, or \$maybe\$. Do not write anything else. Do not explain your answer.
\end{tcolorbox}

\subsection{LongBench}

Due to the nature of the LongBench subsets, there are only per-subset template rather than a common one.

\begin{tcolorbox}[colback=gray!05, colframe=black!40, title=LongBench NarrativeQA]
You are given a story, which can be either a novel or a movie script, and a question. Answer the question asconcisely as you can, using a single phrase if possible. Do not provide any explanation.

\textbf{Story:} \texttt{\$CONTEXT}

Now, answer the question based on the story asconcisely as you can, using a single phrase if possible. Do not provide any explanation.

\textbf{Question:} \texttt{\$QUESTION}

\textbf{Answer:}
\end{tcolorbox}

\begin{tcolorbox}[colback=gray!05, colframe=black!40, title=LongBench HotpotQA]
Answer the question based on the given passages. Only give me the answer and do not output any other words.

The following are given passages.

\texttt{\$CONTEXT}

Answer the question based on the given passages. Only give me the answer and do not output any other words.

\textbf{Question:} \texttt{\$QUESTION}

\textbf{Answer:}
\end{tcolorbox}

\begin{tcolorbox}[colback=gray!05, colframe=black!40, title=LongBench 2WikiMQA]
Answer the question based on the given passages. Only give me the answer and do not output any other words.

The following are given passages.

\texttt{\$CONTEXT}

Answer the question based on the given passages. Only give me the answer and do not output any other words.

\textbf{Question:} \texttt{\$QUESTION}

\textbf{Answer:}
\end{tcolorbox}

\begin{tcolorbox}[colback=gray!05, colframe=black!40, title=LongBench TriviaQA]
Answer the question based on the given passage. Only give me the answer and do not output any other words. The following are some examples.

\texttt{\$CONTEXT}

\texttt{\$QUESTION}
\end{tcolorbox}